\documentclass[runningheads]{llncs}
\usepackage[T1]{fontenc}
\usepackage{graphicx}
\usepackage{amssymb}            
\usepackage{mathtools}          

\usepackage{algorithm}
\usepackage{algorithmic}

\usepackage{tikz}

\usepackage{todonotes}
\usepackage{xcolor}

\begin{document}

\title{VDSC: Enhancing Exploration Timing with Value Discrepancy and State Counts}

\titlerunning{VDSC: Enhancing Exploration Timing with Value Discrepancy and State Counts}

 \author{Marius Captari\inst{1}\and
 Remo Sasso\inst{2}\and
 Matthia Sabatelli\inst{1}}
 \authorrunning{M. Captari et al.}
 %
 \institute{University of Groningen, Broerstraat 5, 9712 CP Groningen, Netherlands
 \email{m.c.captari@student.rug.nl}\\
 \email{m.sabatelli@rug.nl}\\
 \and
 Queen Mary University of London, Mile End Road, London E1 4NS, UK\\
 \email{r.sasso@qmul.ac.uk}}

\maketitle 

\begin{abstract}
Despite the considerable attention given to the questions of \textit{how much} and \textit{how to} explore in deep reinforcement learning, the investigation into \textit{when} to explore remains relatively less researched. While more sophisticated exploration strategies can excel in specific, often sparse reward environments, existing simpler approaches, such as $\epsilon$-greedy, persist in outperforming them across a broader spectrum of domains. The appeal of these simpler strategies lies in their ease of implementation and generality across a wide range of domains. The downside is that these methods are essentially a blind switching mechanism, which completely disregards the agent's internal state. In this paper, we propose to leverage the agent's internal state to decide \textit{when} to explore, addressing the shortcomings of blind switching mechanisms. We present Value Discrepancy and State Counts through homeostasis (VDSC), a novel approach for efficient exploration timing. Experimental results on the Atari suite demonstrate the superiority of our strategy over traditional methods such as $\epsilon$-greedy and Boltzmann, as well as more sophisticated techniques like Noisy Nets.


\end{abstract}

\section{Introduction}
\label{sec:introduction}

Despite remarkable successes in deep reinforcement learning (DRL), the core challenges of reinforcement learning (RL) are still restricting the applicability to real-world scenarios due to severe data-inefficiency. Central to these challenges is the exploration-exploitation trade-off \cite{sutton1998introduction}. Every agent encounters this dilemma: should it rely on its current knowledge and take actions that are known to yield good rewards (exploitation), or should it try new actions, potentially discovering better strategies but risking lower immediate rewards (exploration)?

In the domain of DRL, considerable effort has been directed towards understanding the optimal balance of exploration (\textit{how much}) and devising techniques for executing exploratory actions (\textit{how}). This encompasses studies on modulating the exploration rate over time and optimizing action selection strategies for better learning outcomes \cite{auer2002,tokic2010adaptive,osband2016deep}. A rich body of work explores various strategies, such as stochastic action selection, optimistic initialization, and intrinsic motivation \cite{pathak2017curiosity,burda2018large,bellemare2016unifying}.

\subsubsection{Contribution.} In this work, we primarily focus on the lesser researched, yet equally important, \textit{when} aspect of exploration. That is, we are interested in utilizing internal information signals for 'triggering' the agent to start exploring. Introducing Value Discrepancy and State Counts (VDSC) as key information triggers, we propose monitoring and combining them through a homeostasis mechanism. The first trigger, Value Promise Discrepancy (VPD), is inspired by insights from \cite{pislar2021should}. VPD is an online measure that initiates exploratory actions by evaluating the discrepancy between an agent's prior value estimate of a state and the actual cumulative reward received over a certain period. This assessment helps agents gauge the accuracy of their internal value predictions, informing their decision to explore further or exploit current knowledge. However, this trigger may not adequately address issues like sparser reward environments, highly stochastic states, and perhaps most importantly, information regarding the novelty of a state. To remedy this, we introduce a second trigger for a richer exploration behavior. This second trigger involves mapping states to hash codes using the SimHash algorithm \cite{charikar2002similarity}, allowing for the counting of individual hash occurrences \cite{tang2017exploration}. The counts are used to encourage the visitation of novel states, rather than frequently visited states. However, unlike the original work, where an exploration bonus is added as an external reward, we propose tracking this metric as an additional trigger for exploration, preserving the original reward structure and avoiding issues like reward distortion and novelty loops.

\subsubsection{Paper Structure.} The remainder of this paper is structured as follows. Section \ref{sec:preliminaries} provides the foundational background and concepts essential to our study. Section \ref{sec:related_work} presents an overview of a literature review of some of the main ideas relating to exploration in DRL. Section \ref{sec:methods} offers an overview of the two main triggers being used, highlights potential limitations, and presents a way in which both triggers are combined in VDSC. Section \ref{sec:experiments} presents the experiments and the results obtained across various games from the Atari benchmark. Lastly, Section \ref{sec:discussion_and_conclusions} provides concluding remarks, summarizes key insights, and suggests potential directions for future research.

\section{Preliminaries}
\label{sec:preliminaries}

We formulate the RL problem as a Markov Decision Processes (MDP) formally represented as a tuple $\langle\mathcal{S}, \mathcal{A}, \mathcal{R}, \mathcal{P}, \gamma\rangle$. where $\mathcal{S}$ denotes the state space, $\mathcal{A}$ represents the action space, $\mathcal{P}: \mathcal{S} \times \mathcal{A} \times \mathcal{S} \rightarrow [0, 1]$ is the transition function indicating the probability of moving from state $s \in \mathcal{S}$ to state $s'$ in $\mathcal{S}$  given action  $a \in \mathcal{A}$, $ \mathcal{R}: S \times A \rightarrow \mathbb{R} $ is the reward function assigning a real-valued reward $r$ at time-step $t$ to each state-action pair and $ \gamma \in (0,1]$ is the discount factor balancing the importance of immediate versus future rewards. The agent-environment interaction is governed by the agent's policy $\pi: \mathcal{S} \rightarrow \mathcal{A}$ which allows us to define the value of a state $s$ via the state-value function $V^{\pi}(s)$:

\begin{equation*}
V^\pi(s) = \mathbb{E}\Bigg[\sum_{k=0}^{\infty} \gamma^k r_{t+k} | s_t = s,\pi\Bigg],
\end{equation*}

and the value of a state-action pair via the state-action value function 
\begin{equation*}
Q^\pi(s, a) = \mathbb{E}\Bigg[\sum_{k=0}^{\infty} \gamma^k r_{t+k} | s_t = s, a_t = a,\pi\Bigg].
\end{equation*}

Both value functions play a crucial role when it comes to the exploration-exploitation dilemma. The answer to whether it is better to explore or exploit strongly depends on how precise the learned estimates of these value functions are. In fact, in RL we are interested in finding a policy which maximizes the aforementioned value functions, as this allows us to define the optimal policy $\pi^\star$, the policy that realizes the optimal expected return $V^\star(s) = \max_\pi  V^\pi(s), \text{ for all } s \in \mathcal{S}$, and the optimal state-action function $Q^\star(s,a) = \max_\pi  Q^\pi(s,a), \text{ for all } s \in \mathcal{S} \text{ and } a \in \mathcal{A}$. It is well-known that if an optimal value function is learned, e.g. $Q^{\star}(s,a)$, there is no need for exploration, as the optimal policy can be easily derived as $\pi^{\star}(s) = \underset{a\in\mathcal{A}}{\text{argmax}} \ Q^{\star}(s,a) \ \text{for all} \ s \in \mathcal{S}$. 


\section{Related Work}
\label{sec:related_work}

Despite the theoretical foundations of exploration methods being well-documented in finite state-action spaces, with established sample complexity bounds \cite{dann2014policy,azar2017}, their practical applicability diminishes significantly when confronted with large or continuous state spaces. These theoretical constraints often stem from a focus on worst-case scenarios, rendering them less effective when addressing real-world problems. Within DRL, hard exploration problems pose significant challenges. In environments characterized by sparse rewards, conventional random exploration methods struggle to reach successful states. Games like \textsc{Pitfall!} and \textsc{Montezuma’s Revenge} from the Atari-57 benchmark exemplify such hard exploration games \cite{bellemare2016unifying}, sparking the development of more deliberate exploration strategies.

\subsubsection{Intrinsic reward} methods are an emerging strategy that incentivizes diverse state visits and behaviors by introducing additional rewards \cite{schmidhuber2010formal}. Prediction error methods, for instance, compute intrinsic rewards based on prediction model errors when revisiting previously encountered states \cite{pathak2017curiosity,burda2018exploration}. While these methods yield substantial performance gains in hard exploration games, their theoretical underpinnings remain relatively weaker. Furthermore, the introduction of unpredictable random noise, such as the "Noisy-TV" problem \cite{burda2018exploration}, poses challenges for intrinsic motivation methods, potentially leading agents into novelty loops that impede task completion.

\subsubsection{Count-based} methods offer another avenue for encouraging exploration by rewarding states with low visit counts. However, directly counting states becomes impractical in high-dimensional or continuous state spaces. To address this, density models estimate pseudo-counts to approximate state visit frequencies \cite{bellemare2016unifying,zhao2020curiositydriven}. Hashing techniques, such as SimHash \cite{charikar2002similarity} or autoencoders, provide alternative approaches to handle high-dimensional state spaces efficiently \cite{tang2017exploration}.

\subsubsection{Probabilistic} methods offer an alternative by modeling uncertainty over transition, reward, or value functions to guide exploration. Randomized value functions \cite{osband2016deep,osband2018randomized,osband2019deep}, inspired by Thompson sampling, rely on maintaining a distribution that represents uncertainty over the optimal value function through an ensemble of Q-functions to guide exploration by periodically sampling and greedily following a value function. \cite{sasso2023posterior} show that modeling such uncertainty over the transition and reward function through Bayesian Linear Regression instead allows for substantially better exploration and sample efficiency. 

\subsubsection{Uncertainty} based approaches such as uncertainty Bellman equations, which connect the uncertainty from the current time-step to the expected uncertainties at subsequent time-steps \cite{o2018uncertainty}, and Noisy Nets \cite{fortunato2018noisy}, which adds stochasticity through parametric noise to the neural network weights, integrate learning and exploration within a unified estimation framework, offering alternative strategies for efficient exploration. Central to this paper is Value Promise Discrepancy, which measures uncertainty by comparing value function predictions with cumulative reward \cite{pislar2021should}. 

\subsubsection{Goal-based} exploration methods like go-explore \cite{ecoffet2021goexplore} operate by revisiting previously encountered promising states and subsequently exploring from there, using a reward state as a goal. This approach to exploration empowered agents to achieve scores previously deemed unattainable in challenging games such as \textsc{Montezuma’s Revenge}, all without any domain knowledge.

\subsubsection{Dithering} methods such as $\epsilon$-greedy, despite all these advancements in exploration techniques, remain prevalent in practice due to their ease of implementation and full coverage of the state space. However, a notable issue with this strategy is its limited capacity for persistent exploration, hindering significant deviations from its current trajectory. Motivated by this limitation, $\epsilon z$-greedy \cite{dabney2020temporally} preserves the simplicity of $\epsilon$-greedy while enabling deviations through a mechanism that repeats the last action for a random duration sampled from a long-tailed distribution.

\paragraph{} The methods discussed here represent only a fraction of the available approaches, offering insight into the diverse ideas fueling exploration within the field. Our work falls within a subset that focuses on using informed triggers to guide the timing of exploration.

\section{Methods}
\label{sec:methods}

In this section, we start by introducing two existing measurements that can be used to steer exploration as triggers. Trigger-based exploration involves utilizing specific indicators to determine when to explore. These triggers can be categorized as either blind, relying on predefined timers or probabilistic mechanisms like $\epsilon$-greedy, or informed, considering internal states relevant to the agent. Motivated by the limitations of these two metrics, we propose a method to overcome them by combining both signals and dynamically setting a threshold for triggering exploration through a homeostasis mechanism.

\subsection{Value Promise Discrepancy}

One measurement that can effectively serve as a trigger for guiding exploration is the Value Promise Discrepancy (VPD) \cite{pislar2021should}. VPD is an online discrepancy measure between the value function's predicted cumulative reward $k$ steps ago for state $s_{t-k}$ to the actual rewards $r_{t-k, \ldots, t}$ obtained in that time frame. Naturally, the VPD is greater in areas that were infrequently visited by the agent and therefore indicating a lack of exploration. Formally, the VPD over a time interval from $ t-k $ to $ t $ is defined as:

\begin{equation*}
\label{eq:vpd}
VPD(t-k,t) := V(s_{t-k}) - \sum_{i=0}^{k-1} \gamma^i r_{t-i} - \gamma^k V(s_t),
\end{equation*}

where, as introduced in Section \ref{sec:preliminaries}, $V(s)$ denotes the agent's value estimate at time-step $t$ for state $s$, reward $r$ and discount factor $\gamma$. By investigating the difference between expected and achieved values, one can gain insights into areas where the agent's learning may be lagging or where it may be over-optimistic about specific actions or states. This measurement can serve as a proxy for the agent's uncertainty, where a higher uncertainty value should encourage the agent towards more exploration.

While the presented signals offer valuable insights into exploration, their usage as triggers introduces several considerations. For instance, a high VPD may suggest heightened uncertainty, yet this interpretation may oversimplify the situation. Elevated VPD values could merely indicate the agent's early learning stage, where prediction errors are inherently high due to initial value estimates deviating from optimality. Similarly, certain state spaces may naturally exhibit higher volatility or unpredictability, resulting in elevated VPD values. Moreover, VPD fails to capture nuances such as state novelty, a metric that, as reviewed in Section \ref{sec:related_work}, is crucial in exploration. Consequently, relying solely on VPD might obscure underlying reasons for prediction errors and overlook other crucial information.

\subsection{Count-Based Exploration and SimHash}
\label{sec:count_based}

Count-based exploration is a foundational method in RL, leveraging the frequency of state visits to guide the agent's exploratory behavior. The idea behind this approach is that states visited less frequently may offer greater informational or reward opportunities, making them prime candidates for exploration. This approach has seen varied applications, from simpler, discrete state spaces to more complex and high-dimensional environments \cite{strehl2008analysis,tang2017exploration,parisi2022long}. However, as most DRL agents typically operate in high-dimensional spaces, traditional tabular methods for tracking state visits are infeasible. Hence, approximation techniques, like hashing, are often employed to achieve this goal \cite{bellemare2016unifying}.

\begin{figure}[ht]
    \centering
    \begin{minipage}[c]{0.3\textwidth}
        \centering
        \includegraphics[width=0.9\textwidth]{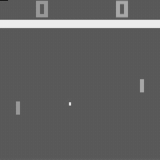}
    \end{minipage}
    \begin{minipage}[c]{0.2\textwidth}
        \centering
        \begin{tikzpicture}
        \draw[->, thick] (0,0) -- (2,0) node[midway, above] {\textbf{Hashing}} node[midway, below] {$k=64$ bits};
        \end{tikzpicture}
    \end{minipage}
    \begin{minipage}[c]{0.3\textwidth}
        \centering
        \begin{tikzpicture}[scale=0.4]
            \foreach \i in {1,2,...,8} {
                \foreach \j in {1,2,...,8} {
                    \pgfmathparse{int(rand*2)}
                    \ifnum\pgfmathresult=0
                        \def\mycolor{white}
                        \def\myvalue{-1}
                    \else
                        \def\mycolor{gray!60}
                        \def\myvalue{1}
                    \fi
                    \draw[fill=\mycolor] (\i,\j) rectangle (\i+1,\j+1);
                    \node at (\i+0.5, \j+0.5) {\myvalue};
                }
            }
        \end{tikzpicture}
    \end{minipage}
    \caption{Possible state hash conversion using $\kappa=64$ bits on an 8$\times$8 grid. Transition from a pre-processed Atari \textsc{Pong} game state (left) to the corresponding hashed state (right).}
\label{fig:hashed_state_example}
\end{figure}
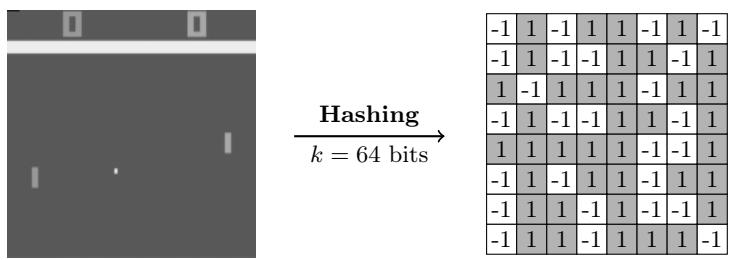

The SimHash function encodes a state $ s \in S $ as $\phi(s) = \text{sgn}(A \cdot g(s))$, where $ A $ is a $ \kappa \times D $ matrix with entries from $ \mathcal{N}(0, 1) $, and $ g: S \rightarrow \mathbb{R}^D $ is an optional pre-processing function. The parameter $ \kappa $, essentially the number of bits of the compressed state, influences collision likelihood and state differentiation. The hashing process can be visualized in Figure \ref{fig:hashed_state_example}.

The counts from the hash table can then be used to calculate a reward bonus, which is added to the agent's final reward at each time step. The reward bonus $r^+$ is calculated according to the following formula:

\begin{equation*}
\label{eq:state-novelty}
    r^+(s) = \frac{\beta}{\sqrt{n(\phi(s))}},
\end{equation*}

where $\beta \in \mathbb{R}_{>0}$ and $\phi$ is the chosen appropriate hash function. The agent is then trained with the final reward $r$ returned by the environment plus the exploration bonus $r^+$. The hashing mechanism inherently clusters similar states, providing a more generalized count and thereby smoothing the exploration landscape. Moreover, the hashing function can be tailored to strike a delicate balance between generalization across states and distinguishing between them.

Count-based approaches, such as the one just presented, also present challenges and drawbacks. Directly augmenting enviroment rewards with bonuses can distort the true environmental feedback, potentially impeding learning or leading to sub-optimal policies \cite{burda2018exploration}. Another concern pertains to the utilization of an additional hyperparameter denoted as $\beta$, which scales the provided bonus reward. Conventionally, a rule of thumb suggests employing a reward bonus with an inverse-square-root dependence. However, research indicates that adopting an inversely proportional relation can lead to faster learning rates \cite{ménard2021fast}. The scaling of this exploration bonus remains unresolved within the context of DRL, representing another parameter requiring fine-tuning.

\subsection{Homeostasis}



One challenge in determining when to explore based on incoming signals is setting a threshold beyond which an exploratory action is triggered. In practice, such signals can vary significantly not only across different environments but also during training. To address this challenge, we present an extended version of the homeostasis mechanism \cite{turrigiano2004homeostatic}, as introduced in \cite{pislar2021should}, which we refer to as 'unified homeostasis.'
The homeostasis mechanism aims to convert a series of numeric signals $x_t$ into binary decisions $y_t \in {0,1}$, aiming for an average switching rate close to a target rate $\rho$. Contextualizing this for our exploration problem, high values of the tracked metric $x_t$ suggest a higher likelihood of exploring ($y_t = 1$), and the target rate $\rho$ refers to our desired exploration rate (similar to $\epsilon$ in $\epsilon$-greedy). Note that the value of $y_t$ only refers to the decision based on whether or not the agent should explore at timestep $t$, that is, should it act greedily ($y_t = 0$) or should it explore ($y_t = 1$). As we are mainly interested in studying the question of \textit{when} to explore in this work, we set the exploratory action randomly.

Unified homeostasis integrates feedback from multiple signals (triggers) by utilizing their respective moving averages to guide the timing of exploration. During each step of the learning process, the method calculates exploration probabilities for all triggers, averaging them to determine the overall exploration probability $\overline{p}$. Subsequently, it samples the decision of whether or not to explore from a Bernoulli distribution using $\overline{p}$ as the parameter. To enhance adaptability, a time scale of interest $\tau := \min(t, \frac{5}{\rho})$ is established. This time scale acts as a reference, dynamically adjusting the influence of recent observations. Specifically, it controls the rate at which the moving averages update, allowing the mechanism to respond effectively to changes in the environment. By incorporating the time scale into the algorithm, the mechanism can balance between capturing short-term fluctuations and maintaining stability over longer periods. The detailed steps of this unified homeostasis mechanism are elaborated in the provided pseudocode (Algorithm \ref{alg:homeostasis}).

\begin{algorithm}
\caption{Unified Homeostasis}
\label{alg:homeostasis}
\begin{algorithmic}[1]
\REQUIRE target rate $\rho$, trigger types = \{VPD, Exploration Bonus\}
\STATE Initialize mean $\overline{x} \leftarrow 0$ variance$, \overline{x^2} \leftarrow 0$, mean transformed $\overline{x^+} \leftarrow 0$ for each trigger
\FOR{$t \in \{1, ...,T\}$}
    \STATE Set time scale of interest $ \tau \gets \min(t, \frac{5}{\rho}) $
    \STATE Set weight of the latest observation $\alpha \gets \frac{1}{\tau}$
    \FOR{each trigger type $i$}
        \STATE Get next signal value $x_t$
         \STATE $ \text{Update } \overline{x} \gets (1 - \alpha) \overline{x} + \alpha x_t $
        \STATE $ \text{Update } \overline{x^2} \gets (1 - \alpha) \overline{x^2} + \alpha (x_t - \overline{x})^2 $
        \STATE $ \text{Standardize } x_t \gets \frac{x_t - \overline{x}}{\sqrt{\overline{x^2}}} $
        \STATE $ \text{Exponentiate } x^+ \gets \exp(x_t)$
        \STATE $ \text{Update transformed average } \overline{x^+} \gets (1 - \alpha) \overline{x^+} + \alpha x^+$
        \STATE  $ \text{Compute exploration probability } p_i \gets \min(1, \rho \frac{x^+}{\overline{x^+}})$
    \ENDFOR
    \STATE $ \text{Average probabilities } \overline{p} \gets \frac{1}{n} \sum_{i=1}^{n} p_i$
    \STATE $ \text{Sample } y_t \sim \text{Bernoulli}(\overline{p})$
\ENDFOR
\end{algorithmic}
\end{algorithm}

\subsection{Value Discrepancy and State Counts (VDSC)}

Given the challenges inherent to each one of the discussed approaches, our proposed approach integrates the agent’s value state prediction errors with the exploration bonus derived from hashed state counts. This integration aims to overcome the limitations associated with relying solely on either VPD or count-based methods, presenting a novel approach to determining \textit{when} to explore.

Combining all of the aforementioned components together, we can summarize our proposed exploration strategy, named VDSC, as the following integration of methodologies that can enhance the decision-making process of DRL agents: The first component involves calculating the VPD values, which act as a proxy for uncertainty in the agent's predictions. This signal evaluates the discrepancy between the agent's expected rewards and the actual outcomes, providing insight into how well the agent's predictions align with reality. The second element is the computation of the exploration bonus, designed to encourage the exploration of less-visited, and therefore more novel, states. This approach assumes that exploring unfamiliar states can reveal new learning opportunities and potentially rewarding experiences, crucial for the agent's development and adaptability. Finally, the strategy includes a process that translates raw signal values - VPD and exploration bonus - into a concrete exploration decision using a unified homeostasis mechanism. This decision-making process aims to balance the need for exploration with exploitation efficiency, ensuring that the agent's actions are both exploratory and informed. The final pseudocode detailing VDSC exploration is outlined in Algorithm \ref{alg:vpd_simhash}.

By combining these components, our strategy aims to address the exploration-exploitation timing dilemma in DRL more effectively. It leverages uncertainty and state novelty to guide the agent toward a balanced and informed exploration path. This approach is expected to enhance learning performance, particularly in complex environments where traditional exploration methods may prove inadequate.

\begin{algorithm}[ht]
    \caption{VDSC exploration strategy}
    \label{alg:vpd_simhash}
    \begin{algorithmic}[1]
        \REQUIRE VPD history length $k$, SimHash bits $\kappa$, exploration rate $t$
        \STATE Define state preprocessor $g : \mathcal{S} \rightarrow \mathbb{R}^D$
        \STATE Initialize matrix $A \in \mathbb{R}^{\kappa \times D}$ with entries drawn i.i.d from $\mathcal{N}(0, 1)$
        \STATE Initialize hash table with values $n(\cdot) \equiv 0$
        \STATE Initialize unified homeostasis instance $\mathcal{H}$ with target exploration rate $t$
        \FOR{timestep $t \in \{1, ...,T\}$}
        \STATE Get current observed state $s_t$
        \STATE Calculate state hash $\phi(s_t) = \text{sgn}(A \cdot g(s_t))$
        \STATE Calculate exploration bonus $(s_t) = \frac{1}{\sqrt{n(\phi(s_t))}}$
        \STATE Update counts $n(\phi(s_t)) \gets n(\phi(s_t)) + 1$
        \STATE Calculate $VPD(t-k,t) = V(s_{t-k}) - \sum_{i=0}^{k-1} \gamma^i r_{t-i} - \gamma^k V(s_t)$
        \STATE Store current $V(s_t)$ and $r_t$
        \STATE Update $\mathcal{H}$ with $VPD$ and exploration bonus
        \STATE Act according to exploration decision from $\mathcal{H}$
        \ENDFOR
    \end{algorithmic}
\end{algorithm}

\section{Experiments}
\label{sec:experiments}

We will now outline our experimental setup, covering the environment, agent, baselines, and hyperparameters. We evaluate VDSC against baseline methods on a subset of Atari games, presenting both quantitative and qualitative results. Additionally, we analyze the impact of individual trigger signals and the temporal structure in exploratory behavior imposed by VDSC.

\subsection{Experimental Setup}
\label{sec:experimental_setup}

\subsubsection{Environments.} We conduct an experimental comparison between VDSC and alternative exploration methods using a subset of games from the Atari Learning Environment (ALE) \cite{bellemare13arcade}. Our evaluation spans four games: \textsc{Pong}, \textsc{Frostbite}, \textsc{Gravitar}, and \textsc{Freeway}, using 3 seeds per game trained over 100M in game frames. The choice of these games is motivated as follows: the latter three environments are well-known to be classified as hard exploration games \cite{bellemare2016unifying}, whereas \textsc{Pong} serves as a simpler, easy exploration alternative with dense rewards. In Figure \ref{fig:atari_57} we provide additional results across the entire Atari game suite. Standard Atari pre-processing techniques are applied as in \cite{mnih2015human}, with the exception of sticky actions which is set to 0.25 and the disabling of termination on life loss, as suggested in \cite{machado2018revisiting}. The final state observed by the agent is transformed from an initial sequence of $210 \times 160$ RGB images to a stack of four $84 \times 84 \times 4$ grayscale images.

\subsubsection{Agents.} We utilize the state-of-the-art DRL agent Rainbow \cite{hessel2018rainbow}, provided by the Dopamine framework \cite{castro2018dopamine}, as the backbone for our experiments to evaluate the performance of VDSC. Two main considerations drive the selection of the Rainbow agent. Firstly, it introduces an additional exploration strategy baseline, Noisy Nets \cite{fortunato2018noisy}, for comparative analysis. Secondly, Rainbow incorporates the Dueling Architecture \cite{wang2016}, which estimates $V^\star(s)$ alongside the advantages of each individual action $A(s,a)= Q^\star(s,a) - V^\star(s)$. This architectural choice is crucial as it aligns with our focus on the network's value estimates for calculating the VPD at each timestep \footnote{Note that several other alternatives to the Dueling Architecture that estimate $V^{\star}(s)$ exist \cite{sabatelli2020deep}\cite{haarnoja2018soft}, all of these can be easily integrated with our proposed exploration strategy.}.

\subsubsection{Baselines.} To benchmark the effectiveness of VDSC, we compare it against three other exploration strategies: $\epsilon$-greedy, Boltzmann, and Noisy Nets. In $\epsilon$-greedy the agent selects a random action at each time step with a set probability $\epsilon \leq 1$, otherwise the action with the highest Q-value is selected. In contrast, Boltzmann samples an action based on a distribution determined by applying the softmax function to the output layer of the network. Lastly, Noisy Nets introduce stochasticity directly into the network weights for more consistent exploration: $w = \mu + \sigma \odot \varepsilon$, where $w$ is the weight, $\mu$ and $\sigma$ are parameters, and $\varepsilon$ is sample noise.

\subsubsection{Hyperparameters.} To maintain the focus of our work on the timing of exploration, we strive to maintain a fixed overall proportion of exploration steps whenever feasible. This involves setting the same initial values and decay schedules for key parameters such as $\epsilon$ in $\epsilon$-greedy, the temperature parameter $\tau$ in Boltzmann exploration, and the target exploration rate in the homeostasis instance. Throughout all Atari games, these parameters are initialized to the same value of $1$ and follow a linear decay schedule over the first 1 million game frames to a final value of $0.01$. Regarding VDSC, a history length of $k=5$ was used when calculating the VPD, and we set the size of the binary code $\kappa=256$ for compressing the states (based on the recommendation from \cite{tang2017exploration}). For Noisy Nets, we kept the original hyperparameter $\sigma_0 = 0.5$, as in \cite{fortunato2018noisy}. We refer the reader to the Supplementary Materials for complete information on the hyperparameters used.

\subsection{Quantitative Results}

We begin by examining the performance of VDSC compared to the selected baseline methods across the discussed Atari games. Additionally, we conduct an ablation study on the same four Atari games to assess the influence of individual trigger signals and their combined effect.

\subsubsection{Comparison to Baseline Methods.}

Figure \ref{fig:atari_vs_baselines} displays a comparative analysis of VDSC against the chosen baselines. Across harder exploration games, notably \textsc{Frostbite} and \textsc{Gravitar}, VDSC consistently outperforms all other methods. This superiority underscores the efficacy of VDSC in managing exploration timing, particularly leveraging information about the agent's internal state in challenging exploration scenarios.

\subsubsection{Ablation Study.}

\begin{figure}[t]
    \centering
    \includegraphics[width=\textwidth]{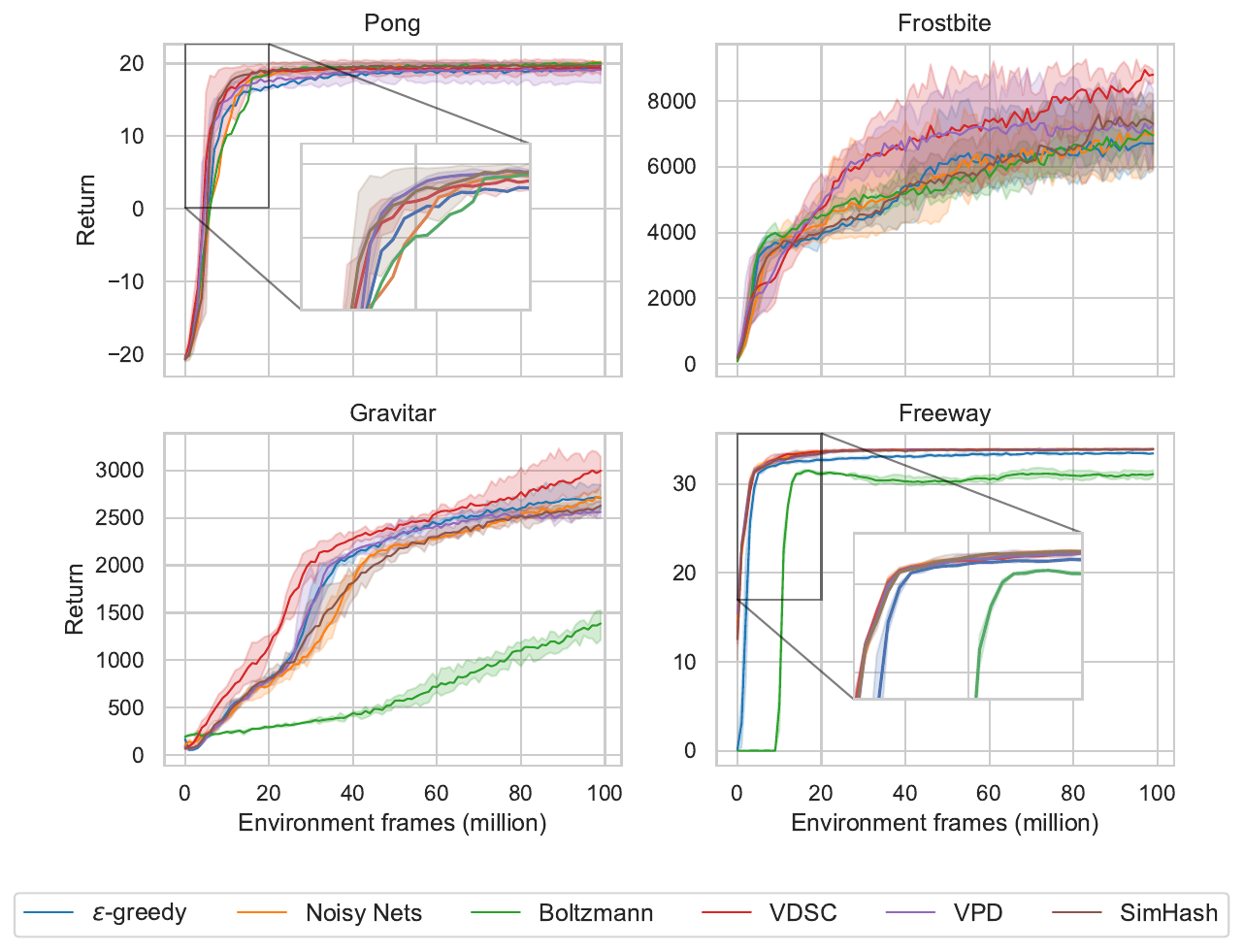}
    \caption{Average episode returns comparing VDSC against baseline methods in addition to each individual trigger in isolation. Shaded regions represent 95\% confidence intervals over 3 random seeds.}
    \label{fig:atari_vs_baselines}
\end{figure}

Figure \ref{fig:atari_vs_baselines} also provides a breakdown of performance by individual signals compared to their combination in VDSC. While each signal demonstrates competitiveness individually against the baseline methods, their synergy in VDSC yields significant performance improvements, especially evident in \textsc{Frostbite} and \textsc{Gravitar}. These findings indicate that the combined signals enable the agent to make more effective exploratory decisions, leading to accelerated learning even with a simplistic combination method.

\begin{figure}[t]
    \centering
    \includegraphics[width=\textwidth]{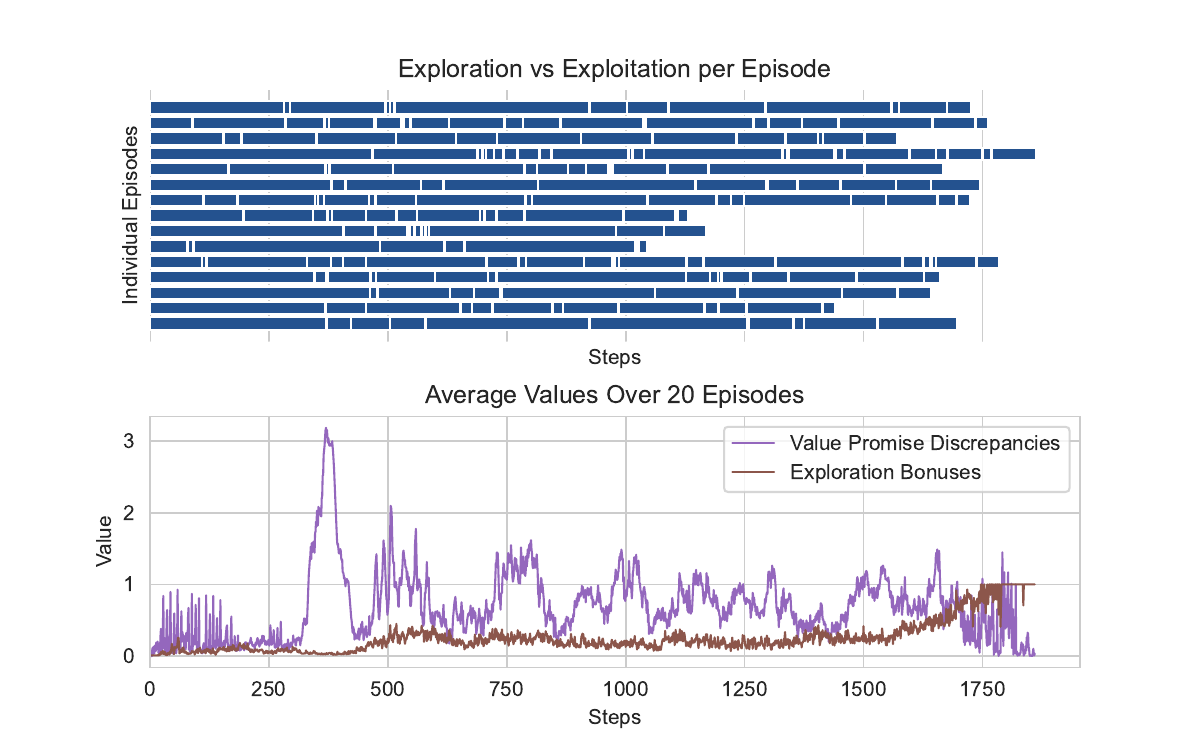}
    \caption{\textbf{Top}: Detailed overview of exploration timings over 20 consecutive training episodes. White vertical bars represent steps in which the agent chose to explore. \textbf{Bottom}: Corresponding average trigger values tracked over the same 20 episodes.}
    \label{fig:atari_exploration} 
\end{figure}

\subsection{Qualitative Results}

Given our study's focus on the timing of exploration, we delve into additional qualitative results to provide deeper insights into VDSC's behavior. Figure \ref{fig:atari_exploration} showcases an analysis conducted over 20 consecutive training episodes using the game \textsc{Frostbite}, captured at approximately 50 million training steps. 

\textsc{Frostbite} presents a game screen with moving ice floes, hazards like birds and fish, and the partially built igloo. The action space is defined by the agent's ability to move in four directions: up, down, left, and right. Points are gained by jumping on ice floes and completing the igloo, but penalties are given if the character falls into the water or collides with enemies. The game is considered a hard exploration environment, given the need for platforming skills and strategic planning to successfully navigate the hazards and build the igloo. Motivated by VDSC results on the game, we aim to investigate the exploratory behavior during training further.

In the top plot of Figure \ref{fig:atari_exploration}, each horizontal bar represents an individual episode. Blue segments denote exploitation phases, while white vertical bars indicate exploratory actions taken by the agent. Examining the detailed temporal structure within episodes (top), a discernible pattern emerges: exploration peaks around the 250-1000 episode step count, with another surge observed towards the episodes' end. This observed behavior aligns with the tracked average values over the same 20 episodes (bottom), where there is a noticeable spike in VPD around the 250-500 step count, followed by a gradual decline throughout the episode durations. Conversely, the average exploration bonus remains relatively low for the majority of episodes, steadily increasing as the episodes near their conclusion. This uptick suggests that the agent encounters more novel states as it approaches losing conditions, prompting more frequent exploration attempts. This strategic increase in exploration endeavors aims to uncover potentially better-performing policies.

It is noteworthy that the overall observed exploration proportion closely matches that of other baselines, such as $\epsilon$-greedy, hovering around 0.01. This observation indicates that a structured and purpose-driven approach to exploration timing, as exemplified by VDSC, can indeed yield enhanced performance, as corroborated by our findings.

\section{Discussion and Conclusions}
\label{sec:discussion_and_conclusions}

We have introduced a novel exploration strategy (VDSC) that combines the agent’s value state prediction errors with counts of the current hashed state, thus balancing long-term uncertainty and state-specific novelty to guide the timing of exploration. The conducted experiments across different environments highlighted the significant role of exploration timing in agent performance and also a significant performance gain over existing traditional exploration methods. Our key contribution is the development of a novel exploration strategy that leverages two distinct signals and a framework for combining them to produce relevant exploration triggers. This approach has demonstrated its potential in complex environments, outperforming traditional methods, particularly where a more refined exploration strategy is required.

This study focuses primarily on value-based methods and their applicability to game environments such as Atari. Extending this exploration strategy to policy gradient methods offers a promising research direction. In policy gradients, exploration is typically governed by stochastic policy updates. By integrating VPD and exploration bonus values into the policy update mechanism, agents can be incentivized to explore more when these values are high, indicating uncertainty or novelty in the environment. This could be achieved by modifying the policy's entropy. An increase in the entropy of the policy when uncertainty is elevated (VPD) or novel states are encountered (high exploration bonus), should encourage the policy to explore more diverse actions. Such modifications would allow the policy to dynamically balance exploration and exploitation based on the uncertainty and novelty of the state space, potentially leading to more robust and efficient learning.

As it was presented in \cite{pislar2021should}, another possible way to gauge the agent's uncertainty is to use an ensemble of networks and measure the discrepancy between them. Furthermore, the usage of a learned domain-specific hash function might lead to further improvements in encoding relevant state information, as it has been shown by \cite{tang2017exploration}.

Lastly, the way in which the signals are combined is also an open question, with some immediate candidates being taking the maximum probability value instead of the average, or perhaps introducing some sort of linear combination which is optimized by an external meta-controller.

\begin{credits}

\end{credits}

%
%
%
\newpage
\bibliographystyle{splncs04}
\bibliography{main}

\begin{thebibliography}{10}
\providecommand{\url}[1]{\texttt{#1}}
\providecommand{\urlprefix}{URL }
\providecommand{\doi}[1]{https://doi.org/#1}

\bibitem{auer2002}
Auer, P.: Using confidence bounds for exploitation-exploration trade-offs. In: Journal of Machine Learning Research. pp. 397--422 (2002)

\bibitem{azar2017}
Azar, M.G., Osband, I., Munos, R.: Minimax regret bounds for reinforcement learning. In: International Conference on Machine Learning. pp. 263--272. PMLR (2017)

\bibitem{bellemare13arcade}
{Bellemare}, M.G., {Naddaf}, Y., {Veness}, J., {Bowling}, M.: The arcade learning environment: An evaluation platform for general agents. Journal of Artificial Intelligence Research  \textbf{47},  253--279 (2013)

\bibitem{bellemare2016unifying}
Bellemare, M., Srinivasan, S., Ostrovski, G., Schaul, T., Saxton, D., Munos, R.: Unifying count-based exploration and intrinsic motivation. Advances in neural information processing systems  \textbf{29} (2016)

\bibitem{burda2018large}
Burda, Y., Edwards, H., Pathak, D., Storkey, A., Darrell, T., Efros, A.A.: Large-scale study of curiosity-driven learning. In: International Conference on Learning Representations (2018)

\bibitem{burda2018exploration}
Burda, Y., Edwards, H., Storkey, A., Klimov, O.: Exploration by random network distillation (2018)

\bibitem{castro2018dopamine}
Castro, P.S., Moitra, S., Gelada, C., Kumar, S., Bellemare, M.G.: Dopamine: A research framework for deep reinforcement learning. arXiv preprint arXiv:1812.06110  (2018)

\bibitem{charikar2002similarity}
Charikar, M.S.: Similarity estimation techniques from rounding algorithms. In: Proceedings of the thiry-fourth annual ACM symposium on Theory of computing. pp. 380--388 (2002)

\bibitem{dabney2020temporally}
Dabney, W., Ostrovski, G., Barreto, A.: Temporally-extended $\varepsilon$-greedy exploration. In: International Conference on Learning Representations (2020)

\bibitem{dann2014policy}
Dann, C., Neumann, G., Peters, J., et~al.: Policy evaluation with temporal differences: A survey and comparison. Journal of Machine Learning Research  \textbf{15},  809--883 (2014)

\bibitem{ecoffet2021goexplore}
Ecoffet, A., Huizinga, J., Lehman, J., Stanley, K.O., Clune, J.: Go-explore: a new approach for hard-exploration problems (2021)

\bibitem{fortunato2018noisy}
Fortunato, M., Azar, M.G., Piot, B., Menick, J., Hessel, M., Osband, I., Graves, A., Mnih, V., Munos, R., Hassabis, D., Pietquin, O., Blundell, C., Legg, S.: Noisy networks for exploration. In: International Conference on Learning Representations (2018)

\bibitem{haarnoja2018soft}
Haarnoja, T., Zhou, A., Abbeel, P., Levine, S.: Soft actor-critic: Off-policy maximum entropy deep reinforcement learning with a stochastic actor. In: International conference on machine learning. pp. 1861--1870. PMLR (2018)

\bibitem{hessel2018rainbow}
Hessel, M., Modayil, J., Van~Hasselt, H., Schaul, T., Ostrovski, G., Dabney, W., Horgan, D., Piot, B., Azar, M., Silver, D.: Rainbow: Combining improvements in deep reinforcement learning. In: Proceedings of the AAAI conference on artificial intelligence. vol.~32 (2018)

\bibitem{machado2018revisiting}
Machado, M.C., Bellemare, M.G., Talvitie, E., Veness, J., Hausknecht, M., Bowling, M.: Revisiting the arcade learning environment: Evaluation protocols and open problems for general agents. Journal of Artificial Intelligence Research  \textbf{61},  523--562 (2018)

\bibitem{ménard2021fast}
M{\'e}nard, P., Domingues, O.D., Jonsson, A., Kaufmann, E., Leurent, E., Valko, M.: Fast active learning for pure exploration in reinforcement learning. In: International Conference on Machine Learning. pp. 7599--7608. PMLR (2021)

\bibitem{mnih2015human}
Mnih, V., Kavukcuoglu, K., Silver, D., Rusu, A.A., Veness, J., Bellemare, M.G., Graves, A., Riedmiller, M., Fidjeland, A.K., Ostrovski, G., et~al.: Human-level control through deep reinforcement learning. nature  \textbf{518}(7540),  529--533 (2015)

\bibitem{osband2018randomized}
Osband, I., Aslanides, J., Cassirer, A.: Randomized prior functions for deep reinforcement learning. Advances in Neural Information Processing Systems  \textbf{31} (2018)

\bibitem{osband2016deep}
Osband, I., Blundell, C., Pritzel, A., Van~Roy, B.: Deep exploration via bootstrapped dqn. Advances in neural information processing systems  \textbf{29} (2016)

\bibitem{osband2019deep}
Osband, I., Van~Roy, B., Russo, D.J., Wen, Z., et~al.: Deep exploration via randomized value functions. J. Mach. Learn. Res.  \textbf{20}(124),  1--62 (2019)

\bibitem{o2018uncertainty}
O’Donoghue, B., Osband, I., Munos, R., Mnih, V.: The uncertainty bellman equation and exploration. In: International Conference on Machine Learning. pp. 3836--3845 (2018)

\bibitem{parisi2022long}
Parisi, S., Tateo, D., Hensel, M., D’eramo, C., Peters, J., Pajarinen, J.: Long-term visitation value for deep exploration in sparse-reward reinforcement learning. Algorithms  \textbf{15}(3), ~81 (2022)

\bibitem{pathak2017curiosity}
Pathak, D., Agrawal, P., Efros, A.A., Darrell, T.: Curiosity-driven exploration by self-supervised prediction. In: International conference on machine learning. pp. 2778--2787. PMLR (2017)

\bibitem{pislar2021should}
Pislar, M., Szepesvari, D., Ostrovski, G., Borsa, D.L., Schaul, T.: When should agents explore? In: International Conference on Learning Representations (2021)

\bibitem{sabatelli2020deep}
Sabatelli, M., Louppe, G., Geurts, P., Wiering, M.A.: The deep quality-value family of deep reinforcement learning algorithms. In: 2020 International Joint Conference on Neural Networks (IJCNN). pp.~1--8. IEEE (2020)

\bibitem{sasso2023posterior}
Sasso, R., Conserva, M., Rauber, P.: Posterior sampling for deep reinforcement learning. In: International Conference on Machine Learning. pp. 30042--30061. PMLR (2023)

\bibitem{schmidhuber2010formal}
Schmidhuber, J.: Formal theory of creativity, fun, and intrinsic motivation (1990--2010). IEEE transactions on autonomous mental development  \textbf{2}(3),  230--247 (2010)

\bibitem{strehl2008analysis}
Strehl, A.L., Littman, M.L.: An analysis of model-based interval estimation for markov decision processes. Journal of Computer and System Sciences  \textbf{74}(8),  1309--1331 (2008)

\bibitem{sutton1998introduction}
Sutton, R.S., Barto, A.G.: Reinforcement Learning: {A}n Introduction. The MIT Press, Cambridge, MA (1998)

\bibitem{tang2017exploration}
Tang, H., Houthooft, R., Foote, D., Stooke, A., Xi~Chen, O., Duan, Y., Schulman, J., DeTurck, F., Abbeel, P.: \# exploration: A study of count-based exploration for deep reinforcement learning. Advances in neural information processing systems  \textbf{30} (2017)

\bibitem{tokic2010adaptive}
Tokić, M.: Adaptive $\epsilon$-greedy exploration in reinforcement learning based on value differences. Annual Conference on Artificial Intelligence pp. 203--210 (2010)

\bibitem{turrigiano2004homeostatic}
Turrigiano, G.G., Nelson, S.B.: Homeostatic plasticity in the developing nervous system. Nature reviews neuroscience  \textbf{5}(2),  97--107 (2004)

\bibitem{wang2016}
Wang, Z., Schaul, T., Hessel, M., Hasselt, H., Lanctot, M., Freitas, N.: Dueling network architectures for deep reinforcement learning. In: International conference on machine learning. pp. 1995--2003. PMLR (2016)

\bibitem{zhao2020curiositydriven}
Zhao, R., Tresp, V.: Curiosity-driven experience prioritization via density estimation (2020)

\end{thebibliography}

\appendix

\newpage

\begin{figure}[ht]
    \centering
    \includegraphics[width=0.88\textwidth]{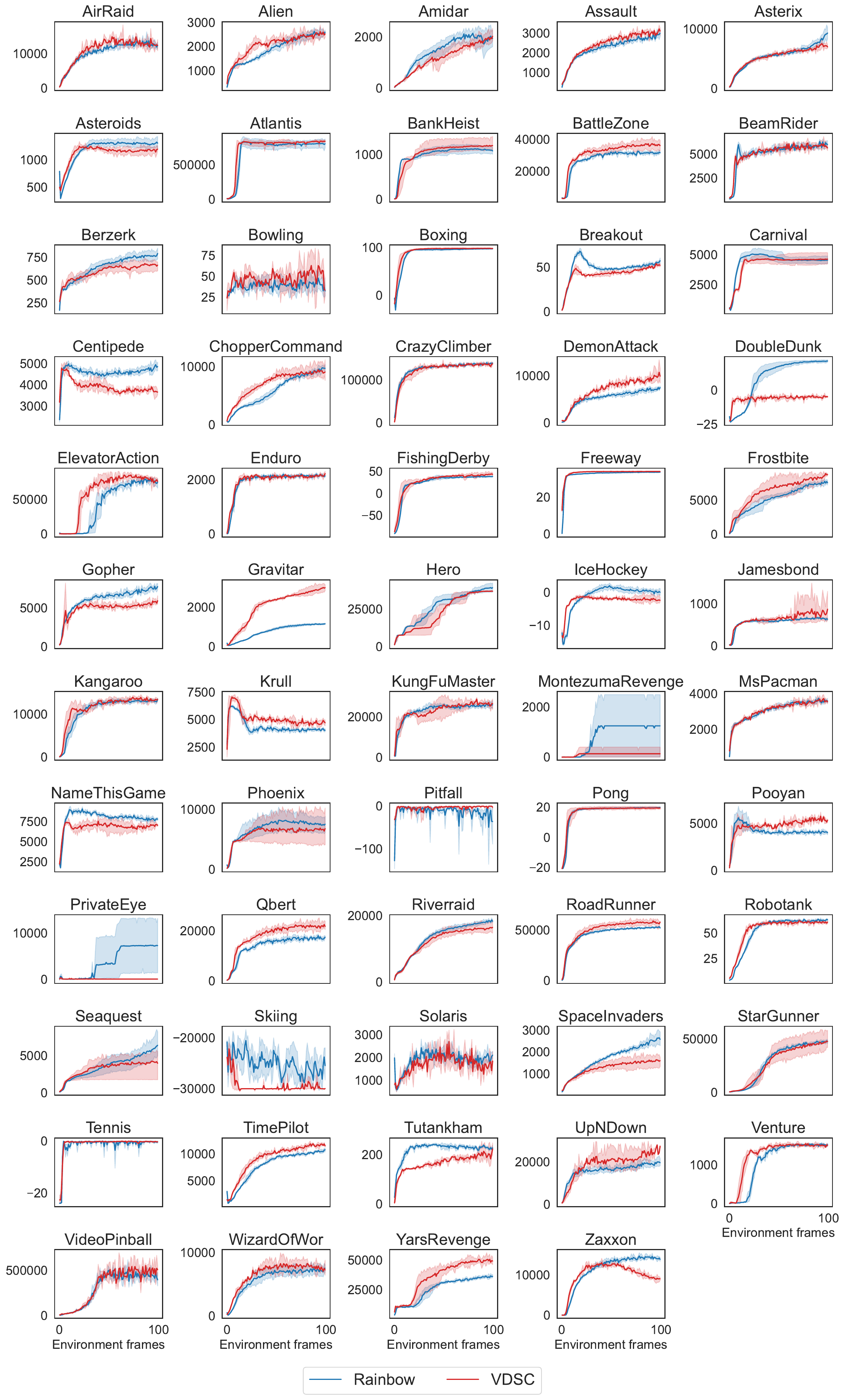}
    \caption{Average episode returns for all Atari games. Shaded regions represent 95\% confidence intervals over 3 random seeds.}
    \label{fig:atari_57}
\end{figure}

\renewcommand{\arraystretch}{0.9}

\end{document}


\title{Supplementary Material}

\titlerunning{Enhancing Exploration Timing with Value Discrepancy and State Counts}


\maketitle 

\section{Atari enviroment}

We adhere to the standard Atari pre-processing protocol, which involves three primary steps: converting RGB frames to grayscale to reduce dimensionality, downsampling to decrease computational load, and stacking consecutive frames to provide motion information. This transforms the original 210x160 RGB images into a stack of four 84x84 grayscale images, allowing the agent to capture essential state information efficiently.

As highlighted in \cite{machado2018revisiting}, different ALE parameter choices can heavily influence the research conclusions, thus a more standard methodology is proposed for evaluation. We follow these recommendations by enabling sticky actions across all tested Atari games. Contrary to the original versions in which the transactions are deterministic, sticky actions make it so that the agent executes its last action, as opposed to the one the agent just selected, with a certain probability $p$. This change aims to reduce the ability of the agent to simply memorize a sequence of actions that can result in a high score, by introducing a form of stochasticity in the transition process. The probability $p$ is set to 0.25, as suggested by \cite{machado2018revisiting}. 

Table \ref{tab:atari-env-hyperparameters} summarizes all the preprocessing hyperparameters used across all the Atari game experiments.

\begin{table}[ht]
    \centering
    \begin{tabular}{c c}
        \hline
        \textbf{Hyperparameter} & \textbf{Value} \\
        \hline
        Grey-scaling & True \\
        Observation down-sampling & (84, 84) \\
        Frames stacked & 4 \\
        Action repetitions & 4 \\
        Reward clipping & [-1, 1] \\
        Terminal on loss of life & False \\
        Sticky actions probability & 0.25 \\
        Max agent steps per episode & 27K \\
        \hline
    \end{tabular}
    \vspace{1em}
    \caption{Preprocessing hyperparameter values for the Atari experimental setup.}
    \label{tab:atari-env-hyperparameters}
\end{table}

\section{Rainbow agent}

Our Rainbow agent utilizes the JAX implementation provided by the Dopamine framework \cite{castro2018dopamine}, with necessary extensions added to compute the VPD and exploration bonus on top of the provided code.

The network used throughout all our experiments is identical to the original introduced by Rainbow \cite{hessel2018rainbow}. It consists of three initial convolutional layers, succeeded by a single hidden layer comprising 512 units. Subsequently, the network branches into the value and advantage streams from the dueling architecture, each containing a hidden layer with 512 units. The output layer matches the number of actions specific to each tested game. Detailed hyperparameters, such as the number of channels, filter sizes, and hidden layer units, are provided in Table \ref{tab:network_hyperparameters_rainbow}.

For an overview of the agent's key hyperparameters governing its behavior, including the discount factor, replay memory parameters, learning rate, and training details, please refer to Table \ref{tab:hyperparameters_rainbow}. Moreover, Table \ref{tab:baseline-exploration-parameters} offers a concise summary of the exploration parameters utilized across all tested exploration strategies.

\begin{table}[ht]
    \centering
    \begin{tabular}{c c c}
        \hline
        \textbf{Hyperparameter} & \textbf{Value} \\
        \hline
        Channels  & 32, 64, 64 \\
        Filter size & 8x8, 4x4, 3x3 \\
        Stride & 4, 2, 1 \\
        Hidden layers & 1 \\
        Hidden units & 512 \\
        Output units & Number of actions \\
        \hline
    \end{tabular}
    \vspace{1em}
    \caption{Network hyperparameter settings using the Rainbow agent for the Atari experimental setup.}
    \label{tab:network_hyperparameters_rainbow}
\end{table}

\begin{table}[ht]
    \centering
    \begin{tabular}{c c c}
        \hline
        \textbf{Hyperparameter} & \textbf{Value} \\
        \hline
        Discount $\gamma$ & 0.99 \\
        Update horizon & 3 \\
        Min replay history & 20k \\
        Update period & 4 \\
        Target update frequency & 8k \\
        Optimizer & Adam \\
        Minibatch size & 32 \\
        \hline
        Distributional number of atoms & 51 \\
        Distributional min/max & ${[-10,10]}$ \\
        \hline
        Learning rate 0.0000625 \\
        Optimizer $\epsilon$ & 0.00015 \\
        Number of iterations & 100 \\
        Training steps per iteration & 250k \\
        Max steps per episode & 27k \\
        \hline
        Replay scheme & Prioritized \\
        Replay capacity & 1M \\
        Replay buffer batch size & 32 \\
        \hline
    \end{tabular}
    \vspace{1em}
    \caption{Hyperparameter settings using the Rainbow agent for the Atari experimental setup.}
    \label{tab:hyperparameters_rainbow}
\end{table}

\begin{table}[ht]
    \centering
    \begin{tabular}{c c c}
        \hline
        \textbf{Exploration parameter} & \textbf{Value} \\
        \hline
        Initial $\epsilon / \tau$/target rate & 1 \\
        Final $\epsilon / \tau$/target rate & 0.01 \\
        Decay period in agent steps & 250k \\
        \hline
        VPD value of $k$ & 5  \\
        SimHash number of bits & 256  \\
        \hline
    \end{tabular}
    \vspace{1em}
    \caption{Exploration parameters used across all exploration strategies.}
    \label{tab:baseline-exploration-parameters}
\end{table}


%
%
%
\bibliographystyle{splncs04}
\bibliography{main}